%% file: main.tex
\documentclass[10pt,twocolumn,letterpaper]{article}

\usepackage[pagenumbers]{cvpr}

\input{preamble}

\definecolor{linkblue}{rgb}{0.21,0.49,0.74}
\usepackage[pagebackref,breaklinks,colorlinks,allcolors=linkblue]{hyperref}

\newcommand{\nickname}{VMEdit}

\title{\nickname{}: Zero-Shot Video Motion Editing via Proxy-to-Real Transfer}

\author{Ziyang Song\quad Ran Zhang\textsuperscript{\dag}\quad Jinxu Xiang\quad Alan Zhao\\
Tencent Video Solar System
}

\hypersetup{
  pdftitle={VMEdit: Zero-Shot Video Motion Editing via Proxy-to-Real Transfer},
  pdfauthor={Ziyang Song, Ran Zhang, Jinxu Xiang, Alan Zhao}
}

\makeatletter
\patchcmd{\@maketitle}{\vskip .375in}{\vskip .2in}{}{%
  \PackageError{\nickname{}}{Unable to adjust the title top spacing}{}%
}
\patchcmd{\@maketitle}{\vspace*{24pt}}{\vspace*{16pt}}{}{%
  \PackageError{\nickname{}}{Unable to adjust title-to-author spacing}{}%
}
\patchcmd{\@maketitle}{\vspace*{12pt}}{\vspace*{2pt}}{}{%
  \PackageError{\nickname{}}{Unable to adjust author-to-figure spacing}{}%
}
\apptocmd{\@maketitle}{\input{fig/1_teaser}}{}{%
  \PackageError{\nickname{}}{Unable to insert the first-page figure}{}%
}
\makeatother

\begin{document}
\maketitle
\begingroup
\renewcommand{\thefootnote}{\fnsymbol{footnote}}
\footnotetext[2]{Corresponding author: Ran Zhang.}
\endgroup
\input{fig/2_overview}
\input{sec/0_abstract}    
\input{sec/1_intro}
\input{sec/2_liter}
\input{sec/3_method}

\input{sec/4_expm}
\input{fig/6_qualitative}
\input{sec/5_sum}

{
    \small
    \bibliographystyle{ieeenat_fullname}
    \bibliography{main}
}

\clearpage
\appendix
\input{sec/6_app}

\end{document}

%% file: preamble.tex
\usepackage{multirow}
\usepackage{stfloats}
\usepackage{flushend}
\usepackage[normalem]{ulem}
\raggedend

%% file: fig/1_teaser.tex
% Included in the full-width title block, above all two-column text.
\begin{minipage}{\textwidth}
  \centering
  \captionsetup{hypcap=false}
  \includegraphics[width=\linewidth]{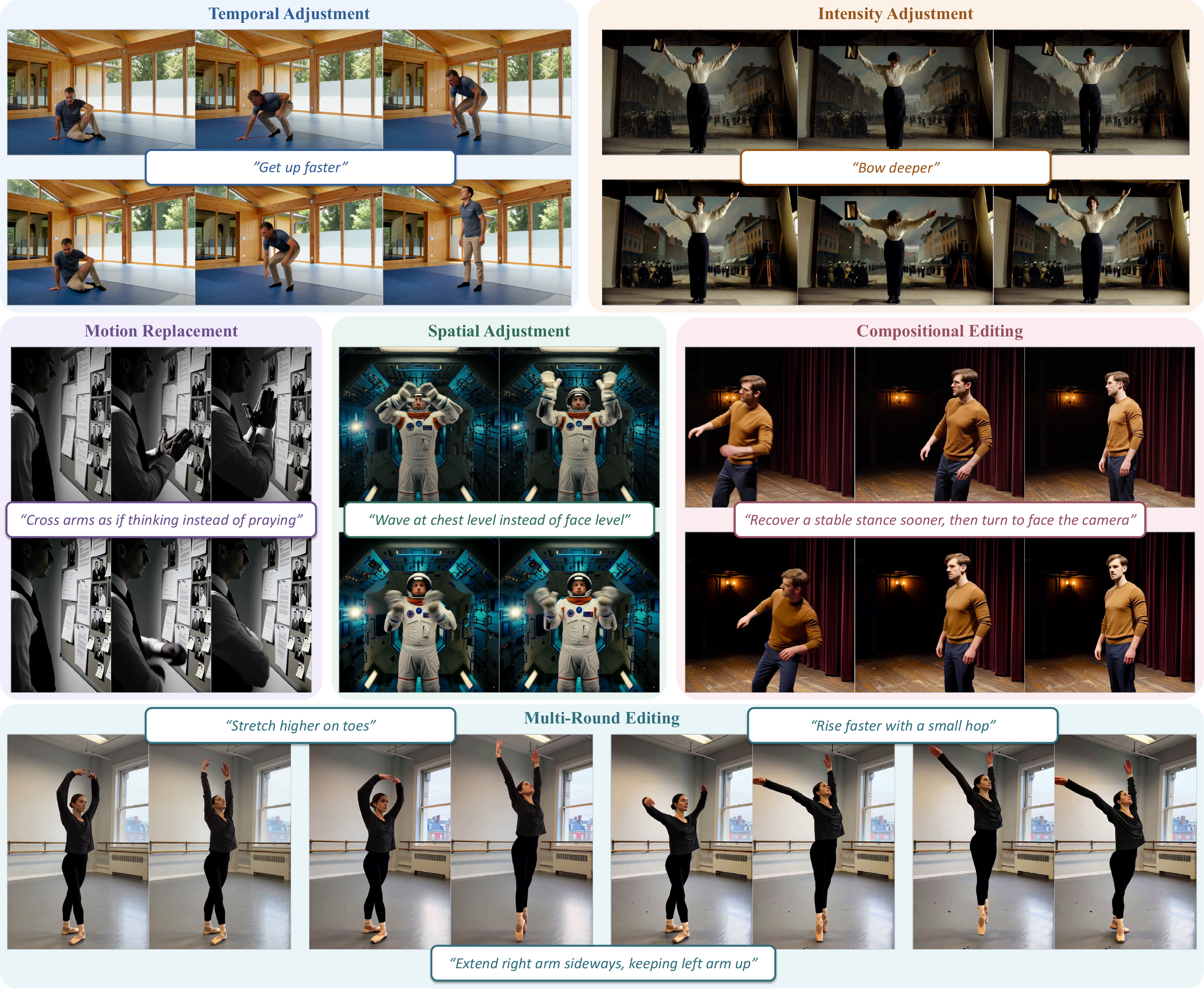}
  \captionof{figure}{\textbf{Instruction-based video motion editing with \nickname{}.} Each example shows the source video above and the edited video below, except for the bottom row which shows the original video followed by three successive editing rounds.}
  \label{fig:teaser}
\end{minipage}
\par\vspace{8pt}

%% file: fig/2_overview.tex
% Queued on page 1 so the double-column float appears at the top of page 2.
\begin{figure*}[t]
  \centering
  \includegraphics[width=\linewidth]{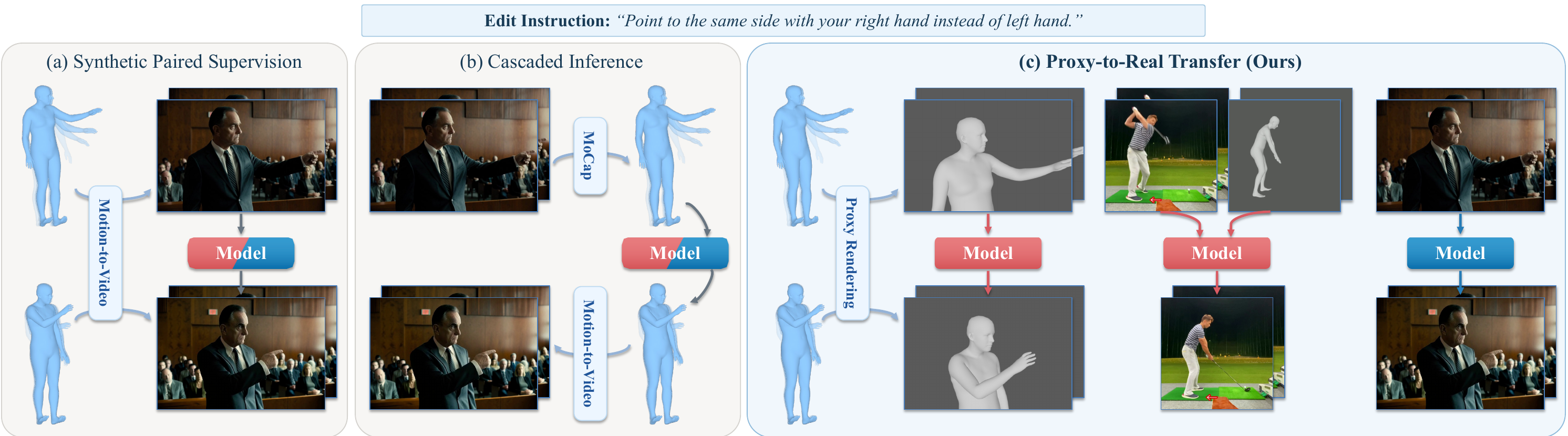}
  \caption{\textbf{Comparison of three paradigms to learn video motion editing from paired 3D motions.} Red and blue indicate models in training and inference respectively, while split red--blue icons denote models used in both stages.}
  \label{fig:overview}
\end{figure*}

%% file: sec/0_abstract.tex
\begin{abstract}
Instruction-based video motion editing provides an intuitive way to apply precise, language-guided modifications to human actions in a video while preserving the original appearance, scene, and unaffected motion. However, this task remains challenging due to the scarcity of paired real-video editing data. To overcome this bottleneck, we present VMEdit, a zero-shot video motion editing framework that requires neither paired real-video supervision nor a cascaded inference pipeline. Our key idea is to learn instruction-based motion editing on coarse visual proxies rendered from accessible 3D motion pairs, while jointly learning an auxiliary motion-guided video generation task that bridges the proxy and real-video domains. This cross-domain alignment enables the instruction-conditioned editing capability learned on proxies to directly transfer to real videos at inference. To construct accurate proxy editing supervision, we further develop a self-improving annotation agent that combines MLLM visual reasoning with 3D kinematic analysis. On our newly established VMEBench, VMEdit significantly outperforms existing methods in motion editing accuracy while maintaining competitive preservation of visual appearance and unaffected motion. Project page: \url{https://szy-young.github.io/vmedit}
\end{abstract}

%% file: sec/1_intro.tex
\section{Introduction}
\label{sec:intro}

Recent advances in video generation and editing have enabled impressive visual content creation and natural-looking edits~\cite{wan21,vace,univideo}. However, existing video editing methods primarily focus on modifying object and character appearances, swapping backgrounds, or adjusting overall style and lighting, without altering the video's spatiotemporal structure~\cite{tokenflow,anyv2v,videopainter,ditto}. Although some studies have explored the editing of character and object motions, they typically require precise and detailed control signals, such as keypoint tracks~\cite{motionv2v,editbytrack} or reference videos~\cite{mtvcraft,wananimate}. For AIGC artists and creators, crafting such intricate control signals is often labor-intensive and time-consuming. In contrast, instruction-based video motion editing offers a more intuitive and user-friendly alternative. As shown in Figure~\ref{fig:teaser}, it uses natural-language prompts to modify specific motion attributes—such as action speed, intensity, position, or semantics—while preserving the character's appearance and unedited motion aspects.

Despite being easy to use, instruction-based video motion editing remains challenging due to the scarcity of paired real training videos. Directly training such a system requires source-target video pairs sharing the same characters, environments, camera views, and most actions, except for local differences in the timing, magnitude, or type of underlying sub-actions. Obtaining such video pairs at scale is extremely difficult. In contrast, 3D motion representations largely abstract away the visual appearance of characters and environments, making it easier to retrieve partially similar motions and characterize their differences. Prior works~\cite{motionfix} demonstrate the feasibility of constructing motion editing datasets by retrieving similar motions from large-scale motion repositories and annotating their differences. This motivates us to explore how paired 3D motions can facilitate motion editing in real-world videos.

A straightforward idea is to synthesize paired training videos from paired 3D motions via motion-guided video generation~\cite{mtvcraft,onetoall,wananimate,scail2}, as in Figure~\ref{fig:overview}(a), which is computationally costly and may mislead model training with synthesis artifacts. An alternative is to learn motion editing on paired 3D motions~\cite{motionfix,simmotionedit,motionrefit} and adopt a multi-stage inference pipeline as in Figure~\ref{fig:overview}(b): 1) estimating 3D motion from the input video, 2) editing the estimated 3D motion via the trained model, and 3) synthesizing the target video conditioned on the edited motion and source appearance~\cite{mtvcraft,wananimate,scail2}. However, this cascaded pipeline suffers from error accumulation from imperfect motion estimation and heavy inference overhead from running multiple models sequentially. These limitations motivate a different solution: Is it possible to directly transfer the editing capability learned from paired 3D motions to real videos at inference?

We answer this question with \textbf{\nickname{}}, a framework for zero-shot video motion editing without requiring paired real training videos or a cascaded inference pipeline. Our key idea is to learn motion editing on a coarse visual proxy constructed from paired 3D motions, while bridging the modality gap between this proxy and real videos for inference-time transfer. As shown in Figure~\ref{fig:overview}(c), we represent 3D motions as rendered videos of textureless human meshes, which can be easily obtained and share a similar spatiotemporal structure with real human videos, despite lacking realistic character and scene appearances. Within a unified model, our motion editing task takes a motion-rendered proxy video and a textual instruction to generate the edited proxy video. Meanwhile, the model jointly learns an auxiliary motion-guided video generation task that takes a reference real video and a target proxy video to synthesize another real video preserving the reference appearance while following the target motion. This auxiliary training task aligns the human body representations between real videos and motion-rendered proxies for seamless proxy-to-real capability transfer, allowing the model to directly edit a source real video from a text instruction at inference.

To learn effective editing capabilities on proxies, high-quality instruction-based supervision is essential. However, human-annotated editing instructions in existing datasets~\cite{motionfix} are labor-intensive and prone to ambiguity and even factual errors. To address this, we develop a self-improving annotation agent that integrates two complementary sources: a multimodal large language model (MLLM) visually compares motion-rendered proxies to capture high-level semantics, while kinematic analysis of the underlying 3D pose sequences grounds the descriptions in measurable kinematic properties. Across annotation rounds, the agent leverages verification uncertainties to iteratively expand its kinematic property set, progressively yielding accurate and verifiable editing instructions.

In summary, our main contributions are threefold:
\begin{itemize}
    \item We propose \textbf{\nickname{}}, a zero-shot video motion editing framework that directly transfers editing capabilities from motion-rendered proxies to real videos without paired real training videos or a cascaded pipeline.
    \item We develop a self-improving annotation agent that combines visual reasoning and 3D kinematic analysis to construct accurate motion-editing supervision.
    \item We establish a comprehensive benchmark and evaluation protocol for instruction-based video motion editing, where \nickname{} achieves superior motion editing accuracy while faithfully preserving non-target video content.
\end{itemize}

%% file: sec/2_liter.tex
\section{Related Work}
\label{sec:related_work}

\subsection{Video Editing}

Video editing has rapidly evolved from task-specific pipelines---relying on feature correspondence, first-frame propagation, inpainting, or synthetic instruction tuning~\cite{tokenflow, anyv2v, videopainter, ditto, easyv2v, viva, flowdirector, nova}---toward unified understanding-and-generation frameworks~\cite{vace, unic, videocof, editverse, reco, omnivideo, univideo, wan21, kling, runway_aleph}. Despite high visual quality, these methods mainly modify appearance (e.g., style, objects, or background) while leaving the source spatiotemporal structure intact. Specifically, synthetic data pipelines typically lift 2D image edits into videos via frame propagation or flow locking~\cite{anyv2v, ditto} with strict spatial alignment, while many architectures bind generation to source attention or features~\cite{tokenflow, flowdirector, nova}, resisting structural pose changes. Recently, several works have explored video motion editing via feature or flow steering~\cite{uniedit, fastvmt} or explicit geometric and reference controls, including 2D/3D point tracks~\cite{motionv2v, editbytrack}, camera trajectories~\cite{recammaster, camdirector}, and driving videos~\cite{mtvcraft, onetoall, imanimation, wananimate}. However, feature steering struggles with fine-grained human articulation, whereas geometric and reference controls require substantial user effort. While EasyV2V~\cite{easyv2v} supports instruction-based motion editing, it focuses on coarse action replacement rather than fine-grained adjustments to speed, magnitude, or position. In contrast, \nickname{} enables fine-grained, instruction-based human motion editing directly from text without external motion controls.

\subsection{3D Motion Editing}

Unlike video editing, 3D motion editing directly manipulates skeletal trajectories or parametric body meshes. MotionFix~\cite{motionfix} pioneers paired source--instruction--target 3D supervision by retrieving and annotating related motions. Following this paradigm, SimMotionEdit~\cite{simmotionedit} adds auxiliary motion-similarity prediction, MotionReFit~\cite{motionrefit} introduces body-part blending (MotionCutMix) with an autoregressive diffusion coordinator, and MotionLab~\cite{motionlab} unifies generation and editing via conditional flow matching. Concurrent studies further advance edit fidelity through cross-axis joint-difference prediction~\cite{crossaxis_motion}, positive--negative invariance learning~\cite{omnisup_motion}, source-preserving denoising~\cite{supermotion}, and multi-human interaction editing~\cite{interedit}. While effective in 3D space, applying these models to real videos requires a cascaded estimation-and-rendering pipeline prone to error accumulation, and existing human-annotated 3D instructions often contain noise. To address both issues, \nickname{} refines paired 3D motion supervision via a self-improving annotation agent and transfers editing capabilities learned on motion-rendered proxies directly to real videos.

%% file: sec/3_method.tex
\input{fig/3_data_pipeline}

\section{\nickname{}}
\label{sec:vmedit}

Given a source video $v_s$ and a textual instruction $c$ describing the desired motion changes, \nickname{} aims to generate a target video $v_t$ with modified human motion while preserving the character's appearance, scene background, and unaffected actions. Direct supervision for this task is difficult to obtain, as it requires paired real videos $(v_s, v_t)$ exhibiting targeted differences in specific motion attributes under otherwise identical visual conditions, together with an accurate description $c$ of those changes.

We address this supervision gap by learning motion editing on coarse visual proxies rendered from paired 3D motions (\ie{}, rendered videos of textureless human meshes), while bridging the modality gap between proxies and real videos through joint training with an auxiliary task. As shown in Figure~\ref{fig:overview}(c), we jointly train a unified model on two complementary tasks. The core \textbf{proxy motion editing} task teaches the model to modify a source motion-rendered proxy $m_s$ into a target proxy $m_t$ according to an editing instruction $c$:
\begin{equation}
\label{eq:task_m2m}
\quad (m_s, c) \longrightarrow m_t.
\end{equation}
Concurrently, an auxiliary \textbf{motion-guided video generation} task teaches the model to synthesize a target real video $v_t$ conditioned on a reference video $v_s$ (providing character and background appearance) and a target motion proxy $m_t$:
\begin{equation}
\label{eq:task_mv2v}
\quad (v_s, m_t) \longrightarrow v_t.
\end{equation}
Aligning human body representations between motion-rendered proxies and real videos thus enables the model to directly transfer the learned editing capability to real videos for zero-shot \textbf{video motion editing} at inference:
\begin{equation}
\label{eq:task_v2v}
\quad (v_s, c) \longrightarrow v_t.
\end{equation}

In the following subsections, we first detail the proxy motion editing task in Sec.~\ref{sec:motion_editing_data} and the auxiliary motion-guided video generation task for proxy-to-real editing capability transfer in Sec.~\ref{sec:motion_guided_video_data}, followed by our unified model architecture and training setup in Sec.~\ref{sec:model_design}.

\definecolor{acceptgreen}{HTML}{1E7E45}
\definecolor{expandbrown}{HTML}{8A5000}
\definecolor{discardred}{HTML}{C43138}

\subsection{Proxy Motion Editing}
\label{sec:motion_editing_data}

\paragraph{Motion-Rendered Visual Proxies.}
To learn the proxy motion editing mapping $(m_s, c) \to m_t$ in Eq.~\eqref{eq:task_m2m}, we collect 5,346 source--target 3D motion pairs from MotionFix~\cite{motionfix}, which are mined from AMASS~\cite{amass} via TMR~\cite{tmr} feature similarity. To make 3D motions directly compatible with the video generation backbone and structurally aligned with real human videos, we convert these 3D sequences into coarse motion-rendered proxy video pairs $(m_s, m_t)$ by animating an untextured SMPL-X~\cite{smplx} body mesh and rendering it against a clean background. This representation is easy to obtain while preserving rich 3D kinematic cues. Furthermore, to improve robustness to diverse camera movements in real videos, each motion pair is rendered under two camera trajectories, applying the same trajectory to both source and target to yield 10,692 proxy pairs.

\paragraph{Self-Improving Annotation Agent.} 
Motion editing supervision requires not only source--target motion pairs but also instructions that accurately describe their differences. However, we find that the raw human annotations in MotionFix~\cite{motionfix} contain ambiguous descriptions and even factual errors, as illustrated in Figure~\ref{fig:data_pipeline} left. Rather than relying on these noisy labels, we discard the raw annotations and develop a self-improving annotation agent to re-annotate the motion pairs from scratch. Our agent grounds descriptions in both visual and kinematic evidence and uses feedback during annotation to continuously improve its analysis capabilities on subsequent samples.

For visual evidence, our agent feeds the motion-rendered proxy pairs $(m_s, m_t)$ into a multimodal large language model (MLLM), \ie{}, GPT 6.1 Sol~\cite{gpt61}, to capture overall motion semantics. From the video pair, the MLLM proposes two types of candidate edits: \textit{observational candidates} for visually apparent changes, and \textit{exploratory hypotheses} for motion attributes worth inspecting given the action semantics. Neither can be finalized without verification, as even advanced MLLMs struggle with fine-grained visual motion perception: camera-relative and body-relative reference frames are easily confused when judging directions, and periodic repetitions are difficult to count under sparse frame sampling. To verify these candidates, the agent incorporates 3D kinematic analysis of the underlying pose sequences as complementary evidence. Specifically, it computes joint-level \textbf{kinematic properties}---such as joint positions, speeds, rotation angles, and trajectory extrema---directly from raw 3D pose sequences, where reference frames are unambiguous and dense temporal sampling enables reliable counting.

Since diverse motion pairs involve a wide range of kinematic properties, it is impractical to predefine an exhaustive set covering every case. We therefore design the agent to progressively expand its kinematic property set during annotation. For each motion pair, the agent computes all properties in its current set and feeds them with $(m_s, m_t)$ to the MLLM to propose and verify candidate edits. At each step, the agent takes one of three actions: 
\textcolor{expandbrown}{\textbf{1) Expand}} the kinematic property set if existing properties are insufficient to verify a candidate edit, in which case the agent implements and computes the required new properties to re-verify the current pair and retains them for future samples.
\textcolor{acceptgreen}{\textbf{2) Accept}} and terminate once all confirmed edit claims are supported by consistent visual and kinematic evidence, key motion differences are covered, and no critical ambiguities remain. 
\textcolor{discardred}{\textbf{3) Discard}} and terminate if key editing relationships remain indeterminate even after property expansion due to inherent motion ambiguity or data quality issues.

The annotation agent successfully annotates 10,692 source--target pairs. As shown in Figure~\ref{fig:data_pipeline} left, the resulting structured annotations replace the noisy raw annotations with accurate and detailed \textit{changed motion} and \textit{preserved motion} descriptions. Beyond re-annotating existing datasets, this pipeline offers a promising path toward scaling up motion-editing supervision: since it requires no manual labels, our agent can be readily combined with feature-similarity-based pair mining~\cite{tmr} to automatically discover and annotate motion pairs from larger-scale 3D motion datasets~\cite{bonesseed}.

\subsection{Auxiliary Motion-Guided Video Generation}
\label{sec:motion_guided_video_data}

While motion editing is learned in the proxy domain, training solely on untextured proxies struggles to reliably transfer this capability and risks degrading visual fidelity in real videos. To bridge this modality gap, we introduce the auxiliary motion-guided video generation task $(v_s, m_t) \to v_t$ in Eq.~\eqref{eq:task_mv2v}, aligning body representations across domains while preserving real video synthesis priors.

To construct training triplets $(v_s, m_t, v_t)$, we require paired real videos $(v_s, v_t)$ sharing consistent subject and scene appearance. Unlike proxy motion editing, which relies on partially similar motion pairs to learn local modifications, this auxiliary task imposes no similarity constraints on the motions in $(v_s, v_t)$. Instead, we exclude pairs with overly similar motions to prevent the model from taking a shortcut by simply copying $v_s$ while ignoring the target motion proxy $m_t$. To collect such pairs at scale, we segment single-shot long videos from an internal collection into shorter clips that naturally share subject and scene appearance while capturing diverse movements (Figure~\ref{fig:data_pipeline} right). Forming candidate pairs $(v_s, v_t)$ from clips within each video, we filter out those with appearance inconsistencies (\eg{}, due to camera panning or severe occlusion) or excessive motion similarity. For each retained pair, we extract the 3D motion of $v_t$ via CameraHMR~\cite{camerahmr} and render it with the untextured SMPL-X template (Sec.~\ref{sec:motion_editing_data}) to obtain $m_t$, yielding 31,123 training triplets.

\subsection{Unified Model Architecture and Training}
\label{sec:model_design}
\label{sec:training_setup}

\paragraph{Unified Generative Architecture.}
We build \nickname{} upon the H3-Ref2V base model~\cite{h3}, which natively accepts multiple reference videos alongside a text instruction, allowing us to unify both training tasks and the target inference task within a single generative architecture. Taking the target inference task $(v_s, c) \to v_t$ in Eq.~\eqref{eq:task_v2v} as an example, during each forward pass, H3-Encoder $\mathcal{E}$ processes the instruction $c$ and sampled frames from $v_s$ to extract multimodal context features $\mathcal{E}(c, v_s)$, while H3-VisualVAE $\Phi$ encodes and patchifies the reference video into visual latent tokens $\Phi(v_s)$. Let $z_\tau$ denote the patchified noisy latent of the target video $v_t$ at flow timestep $\tau$. The context features $\mathcal{E}(c, v_s)$, reference tokens $\Phi(v_s)$, and noisy target tokens $z_\tau$ are concatenated into a unified sequence for the single-stream H3-Omni-Transformer $F_\theta$. Within $F_\theta$, visual tokens are positioned via 3D Rotary Positional Embeddings (RoPE)~\cite{rope} and modulated by timestep-conditioned adaptive layer normalization (AdaLN) to distinguish the noisy target $z_\tau$ from the clean reference conditions. Through joint self-attention, the model predicts the target velocity field:
\begin{equation}
\label{eq:model_forward}
\widehat{u}_\tau = F_\theta\!\left(\mathcal{E}(c, v_s),\, \Phi(v_s),\, z_\tau, \tau\right).
\end{equation}
During generation, the reference tokens remain fixed while the predicted velocity $\widehat{u}_\tau$ iteratively updates $z_\tau$ over successive sampling steps, after which the visual VAE decodes the resulting latent into the output video $\widehat{v}_t$. The two training tasks follow the same forward pass by simply substituting their respective inputs.

\input{fig/4_evaluation}

\paragraph{Guidance-Preserving Flow Matching.}
We fine-tune \nickname{} under the rectified flow framework~\cite{rectifiedflow}. Given a clean target latent $z=\Phi(v_t)$, Gaussian noise $\epsilon\sim\mathcal{N}(0,I)$, and timestep $\tau \in (0, 1)$, the noisy target latent is $z_\tau = \tau z + (1-\tau)\epsilon$ with ground-truth velocity $u = z - \epsilon$. Since H3-Ref2V is classifier-free guidance (CFG) distilled, its predictions reside in a CFG-amplified velocity space. Directly regressing the plain velocity $u$ pulls the model out of this space and degrades its pretrained generative priors. To keep training anchored in the amplified space, we construct a guidance-preserving target $\widetilde{u}$ using the model's no-grad unconditional prediction $u_\varnothing = \operatorname{sg}[F_\theta(\mathcal{E}_\varnothing, \Phi(v_s), z_\tau, \tau)]$ under an empty context $\mathcal{E}_\varnothing$:
\begin{equation}
\label{eq:guidance_distillation_loss}
\mathcal{L}_{\mathrm{GP}} = \mathbb{E}\!\left[\left\|\widehat{u}_\tau - \widetilde{u}\right\|_2^2\right], \quad \widetilde{u} = u_\varnothing + \gamma(u - u_\varnothing).
\end{equation}
Here, $\gamma=4$ is the guidance scale and $\operatorname{sg}[\cdot]$ denotes the stop-gradient operator. Both training tasks optimize $\mathcal{L}_{\mathrm{GP}}$ exclusively over the target tokens $z_\tau$.

\paragraph{Implementation Details.}
We fine-tune the Omni-Transformer $F_\theta$ using 256-rank LoRA~\cite{lora} while freezing the pretrained encoder, VAE, and base weights. The model is trained for 2,500 iterations with a batch size of 32, sampling both training tasks with equal probability. We use AdamW with a constant learning rate of $10^{-5}$ and weight decay of $10^{-3}$.

%% file: fig/3_data_pipeline.tex
\begin{figure*}[t]
  \centering
  \includegraphics[width=\linewidth]{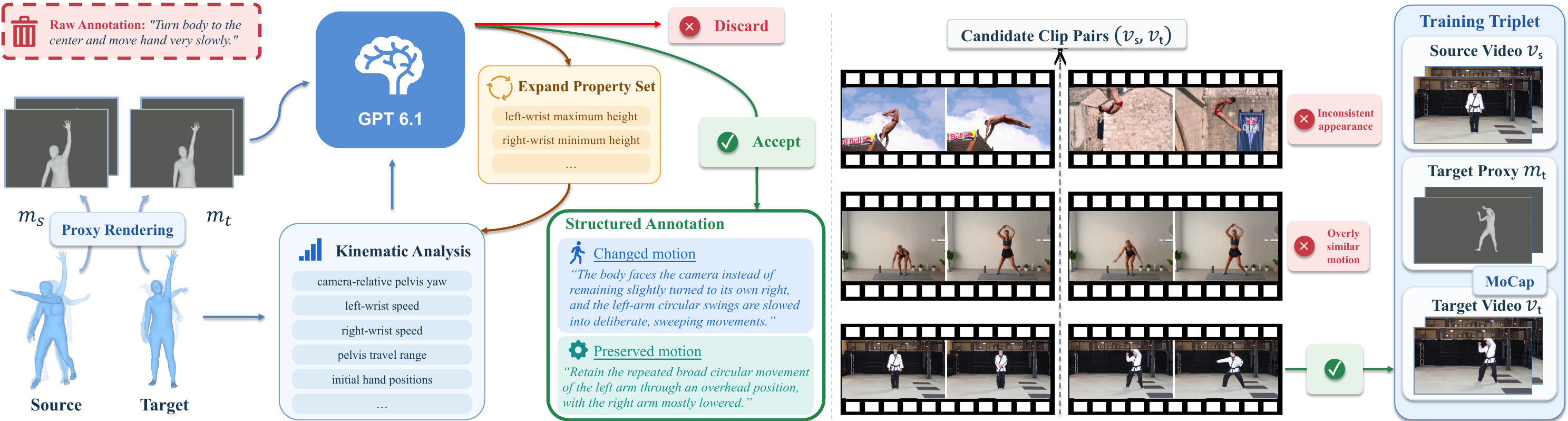}
  \caption{\textbf{Constructing training supervision for \nickname{}.} \textbf{Left - Proxy motion editing:} A self-improving annotation agent fuses visual and kinematic evidence with iterative property set expansion to yield structured editing supervision. \textbf{Right - Motion-guided video generation:} Same-video clip pairs with consistent appearance and distinct motions form cross-modal alignment triplets $(v_s, m_t, v_t)$.}
  \label{fig:data_pipeline}
\end{figure*}

%% file: fig/4_evaluation.tex
\begin{figure}[t]
  \centering
  \includegraphics[width=\linewidth]{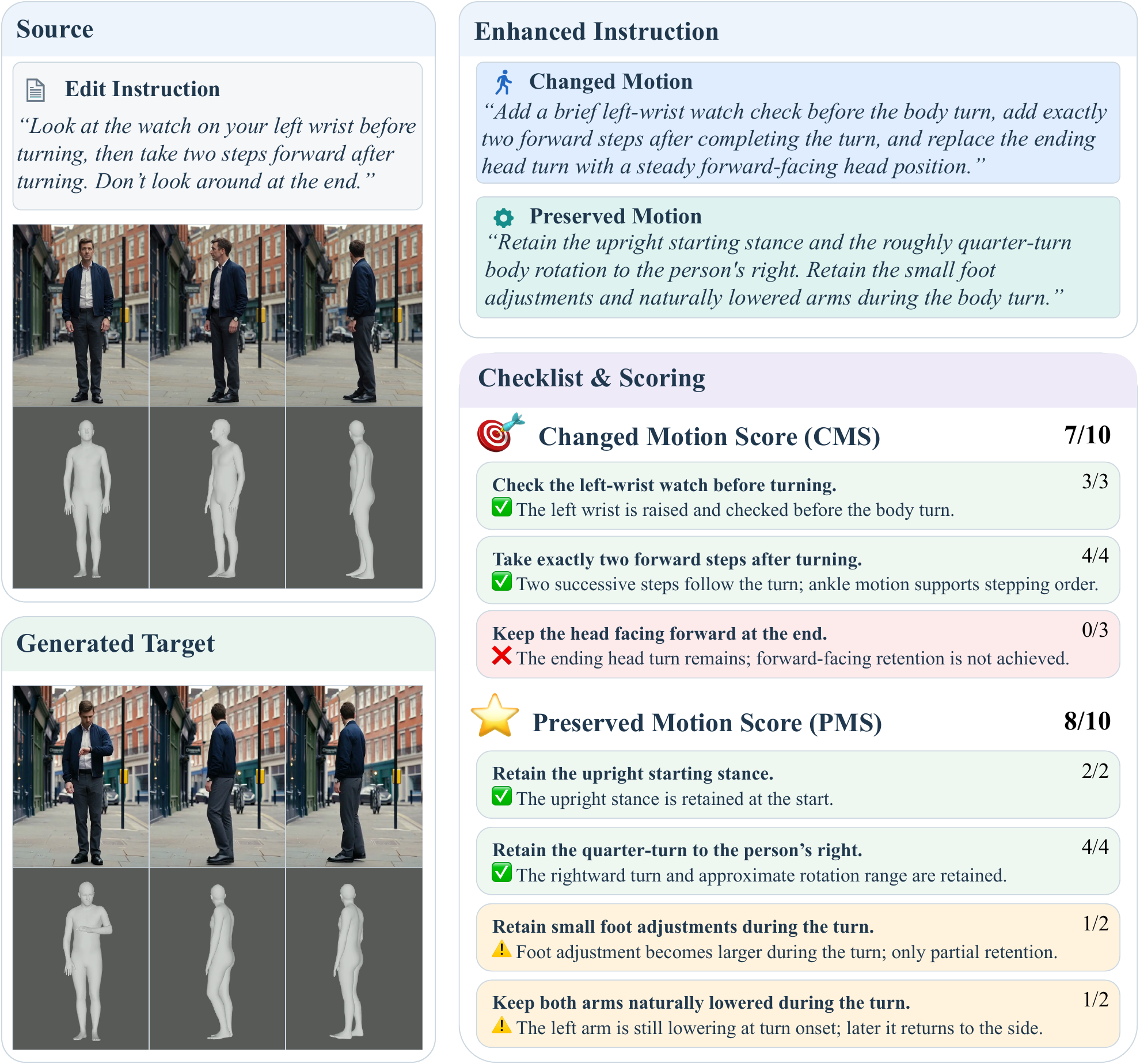}
  \caption{\textbf{An example of CMS and PMS evaluation.}}
  \label{fig:evaluation}
\end{figure}

%% file: sec/4_expm.tex
\section{Experiments}
\label{sec:experiments}

\subsection{VMEBench}
\label{sec:vmebench}

To systematically evaluate instruction-based video motion editing, we construct \textbf{VMEBench} based on the MotionFix test split~\cite{motionfix}. Specifically, we curate representative editing instructions that span diverse modification types---including temporal timing, motion magnitude, action semantics, and their compositions. For each instruction, we design an appropriate scene context and employ H3-T2V~\cite{h3} to synthesize a high-quality source human video depicting the initial action required by the instruction. After careful manual verification and filtering, we retain 52 challenging test samples covering diverse human appearances, backgrounds, and action categories, where the target editing instructions are clearly feasible on the source videos.

\begin{table}[!t]
  \centering
  \caption{\textbf{Quantitative comparison on VMEBench.} H3-Ref2V$^{*}$ is finetuned on our motion-guided video generation task.}
  \label{tab:main_comparison}
  \begingroup
  \fontsize{9}{10.5}\selectfont
  \setlength{\tabcolsep}{2pt}
  \renewcommand{\arraystretch}{1.15}
  \setlength{\arrayrulewidth}{0.3pt}
  \resizebox{\columnwidth}{!}{%
  \begin{tabular}{@{}r|ccc|ccc|cc@{}}
    \toprule
    & \multicolumn{6}{c|}{Instruction Following} & \multicolumn{2}{c}{Appearance} \\
    \cmidrule(l){2-9}
    \multicolumn{1}{c|}{\multirow{2}{*}{Method}} & \multicolumn{3}{c|}{CMS $\uparrow$} & \multicolumn{3}{c|}{PMS $\uparrow$} & \multirow{2}{*}{SC $\uparrow$} & \multirow{2}{*}{BC $\uparrow$} \\
    & {\fontsize{7}{8.5}\selectfont\itshape GPT} & {\fontsize{7}{8.5}\selectfont\itshape Opus} & {\fontsize{7}{8.5}\selectfont\itshape Gemini} & {\fontsize{7}{8.5}\selectfont\itshape GPT} & {\fontsize{7}{8.5}\selectfont\itshape Opus} & {\fontsize{7}{8.5}\selectfont\itshape Gemini} & & \\
    \midrule
    Kling 4.0 Omni~\cite{kling}  & 2.69 & 1.92 & 1.85 & \underline{9.73} & \underline{9.58} & 9.62 & \textbf{0.98} & \textbf{0.97} \\
    \midrule
    UniVideo~\cite{univideo}  & 1.90 & 1.33 & 1.04 & 8.73 & 8.33 & 8.69 & 0.96 & 0.94 \\
    Kiwi-Edit~\cite{kiwiedit}  & 1.19 & 0.67 & 0.46 & \textbf{9.88} & \textbf{9.73} & \textbf{9.83} & \underline{0.97} & \underline{0.95} \\
    H3-Ref2V~\cite{h3}  & 4.75 & 3.98 & 3.98 & 9.67 & 9.50 & 9.67 & 0.95 & 0.94 \\
    \midrule
    {\fontsize{7}{8.5}\selectfont UMO\cite{umo}$+$SCAIL-2\cite{scail2}}  & 5.62 & 4.62 & 4.83 & 6.60 & 5.87 & 6.00 & 0.93 & 0.93 \\
    {\fontsize{7}{8.5}\selectfont UMO\cite{umo}$+$H3-Ref2V\cite{h3}}  & 5.27 & 4.69 & 4.75 & 9.12 & 8.29 & 8.75 & 0.81 & 0.86 \\
    {\fontsize{7}{8.5}\selectfont UMO\cite{umo}$+$H3-Ref2V$^{*}$} & \underline{6.31} & \underline{5.48} & \underline{5.98} & 7.62 & 6.83 & 7.29 & 0.91 & 0.92 \\
    \midrule
    \textbf{\nickname{} (Ours)}  & \textbf{8.00} & \textbf{7.40} & \textbf{8.31} & 9.62 & 9.37 & \underline{9.71} & 0.94 & 0.94 \\
    \bottomrule
  \end{tabular}%
  }
  \endgroup
\end{table}

\subsection{Evaluation Metrics}
\label{sec:eval_metric}

Our evaluation covers two core aspects: 1) \textbf{Instruction following}, which assesses both the accurate execution of requested edits and the faithful retention of unedited motion components; and 2) \textbf{Appearance preservation}, which measures the consistency of subject identity and background scenes with the source video. For appearance preservation, we adopt the DINO and CLIP features from VBench~\cite{vbench} to compute \textbf{Subject Consistency (SC)} and \textbf{Background Consistency (BC)}, respectively. We uniformly sample 16 keyframes from each video, take each target keyframe's maximum similarity across all source keyframes and average these maxima over the target video, thereby accommodating temporal misalignment caused by motion edits. However, evaluating fine-grained instruction following cannot be addressed by off-the-shelf metrics and requires dedicated designs tailored to video motion editing.

To this end, we design an MLLM-based evaluation agent that reports two complementary metrics: \textbf{Changed Motion Score (CMS)}, measuring how accurately target modifications are executed, and \textbf{Preserved Motion Score (PMS)}, evaluating whether unaffected motion characteristics are faithfully retained, as illustrated in Figure~\ref{fig:evaluation}. For each sample, given the source video and edit instruction, the agent first explicates the target changed and preserved motions, and further decomposes them into an itemized checklist with specific score budgets (summing to 10 each for CMS and PMS), which is fixed beforehand and shared across all evaluated methods for fair comparison. During evaluation, we apply CameraHMR~\cite{camerahmr} to extract 3D motions from both the source and generated target videos. Mirroring our annotation agent (Sec.~\ref{sec:motion_editing_data}), we then feed both visual evidence (paired RGB videos and synchronized motion-rendered proxies) and computed 3D kinematic properties into the MLLM to verify and score each checklist item with supporting rationale, summing the item scores to obtain the final CMS and PMS. To mitigate single-model bias, we independently run this evaluation with three distinct MLLMs (GPT 6.1~\cite{gpt61}, Opus 5.5~\cite{opus55}, and Gemini 4~\cite{gemini4}) and report their results separately.

\subsection{Comparison with Prior Works}
\label{sec:comparison}

We compare \nickname{} against three groups of baselines: \textbf{1) Closed-source V2V model}, Kling 4.0 Omni~\cite{kling} (other popular models, \eg{}, Seedance 2.5~\cite{seedance25} and Gemini-Omni-Flash~\cite{geminiomniflash}, are excluded as their human-subject safety filters blocked most human-centric source videos), and \textbf{2) Open-source V2V models}, including UniVideo~\cite{univideo}, Kiwi-Edit~\cite{kiwiedit}, and H3-Ref2V~\cite{h3} (the base model of \nickname{}). Both groups directly edit the source video from text instructions, just like \nickname{} at inference. \textbf{3) Cascaded inference pipelines} first extract 3D source motion via CameraHMR~\cite{camerahmr}, edit it with UMO~\cite{umo}, and synthesize the target video conditioned on the edited motion and source appearance. For the motion-to-video stage, we evaluate SCAIL-2~\cite{scail2}, the base H3-Ref2V~\cite{h3}, and H3-Ref2V$^{*}$ finetuned on our motion-guided video generation task.

As shown in Table~\ref{tab:main_comparison}, \nickname{} effectively accomplishes the specified edits, substantially outperforming all baselines in CMS while competitively preserving non-target content. In contrast, the first two groups of V2V models mostly exhibit poor instruction-based motion editing capability. Their extremely low CMS invalidates their high PMS, as simply duplicating the input video yields a perfect PMS but zero CMS. While the cascaded inference pipelines learn moderate editing capabilities from paired 3D motions, they suffer severely from error accumulation across stages, as evidenced by their significantly lower PMS than the first two groups. Figure~\ref{fig:qualitative_comparison} shows qualitative comparisons, along with successive multi-round edits demonstrated by \nickname{}.

\begin{table}[!t]
  \centering
  \caption{\textbf{Ablation study on VMEBench.}}
  \label{tab:ablation}
  \begingroup
  \fontsize{9}{10.5}\selectfont
  \setlength{\tabcolsep}{2pt}
  \renewcommand{\arraystretch}{1.15}
  \setlength{\arrayrulewidth}{0.3pt}
  \begin{tabular*}{\columnwidth}{@{\extracolsep{\fill}}r|ccc|ccc|cc@{}}
    \toprule
    & \multicolumn{6}{c|}{Instruction Following} & \multicolumn{2}{c}{Appearance} \\
    \cmidrule(l){2-9}
    \multicolumn{1}{c|}{\multirow{2}{*}{Method}} & \multicolumn{3}{c|}{CMS $\uparrow$} & \multicolumn{3}{c|}{PMS $\uparrow$} & \multirow{2}{*}{SC $\uparrow$} & \multirow{2}{*}{BC $\uparrow$} \\
    & {\fontsize{7.5}{9}\selectfont\itshape GPT} & {\fontsize{7.5}{9}\selectfont\itshape Opus} & {\fontsize{7.5}{9}\selectfont\itshape Gemini} & {\fontsize{7.5}{9}\selectfont\itshape GPT} & {\fontsize{7.5}{9}\selectfont\itshape Opus} & {\fontsize{7.5}{9}\selectfont\itshape Gemini} & & \\
    \midrule
    w/o M2M   & 5.40 & 4.65 & 4.96 & \underline{9.65} & \underline{9.42} & 9.65 & \textbf{0.95} & \textbf{0.94} \\
    w/o MV2V   & \underline{7.54} & \underline{7.25} & \underline{7.81} & 9.40 & 9.29 & 9.54 & 0.92 & \underline{0.93} \\
    w/o Anno. Agent   & 5.88 & 5.25 & 5.15 & 9.60 & 9.19 & 9.48 & \underline{0.94} & \textbf{0.94} \\
    \midrule
    \textbf{\nickname{} (Ours)}   & \textbf{8.00} & \textbf{7.40} & \textbf{8.31} & 9.62 & 9.37 & \textbf{9.71} & \underline{0.94} & \textbf{0.94} \\
    \bottomrule
  \end{tabular*}
  \endgroup
\end{table}

\subsection{Ablation Study}
\label{sec:ablation}

\input{fig/5_ablation_mv2v}

Table~\ref{tab:ablation} ablates three key components of our framework: 1) the core proxy motion editing task (M2M), 2) the auxiliary motion-guided video generation task (MV2V), and 3) the self-improving annotation agent (Anno.\ Agent) for fine-grained motion-editing supervision. Evidently, the core M2M task is most indispensable, as CMS drops drastically without it. Data refinement via the annotation agent is also crucial, since training on raw annotations with unrectified ambiguities and factual errors brings only marginal gains in editing capability. Building upon these two pillars, the auxiliary MV2V task enables more reliable transfer of learned editing capabilities from the proxy to the real domain. As illustrated in Figure~\ref{fig:ablation_mv2v}, with the cross-domain alignment from MV2V, motion edits that succeed on the proxy domain carry over more consistently to real videos.

%% file: fig/5_ablation_mv2v.tex
\begin{figure}[t]
  \centering
  \includegraphics[width=\linewidth]{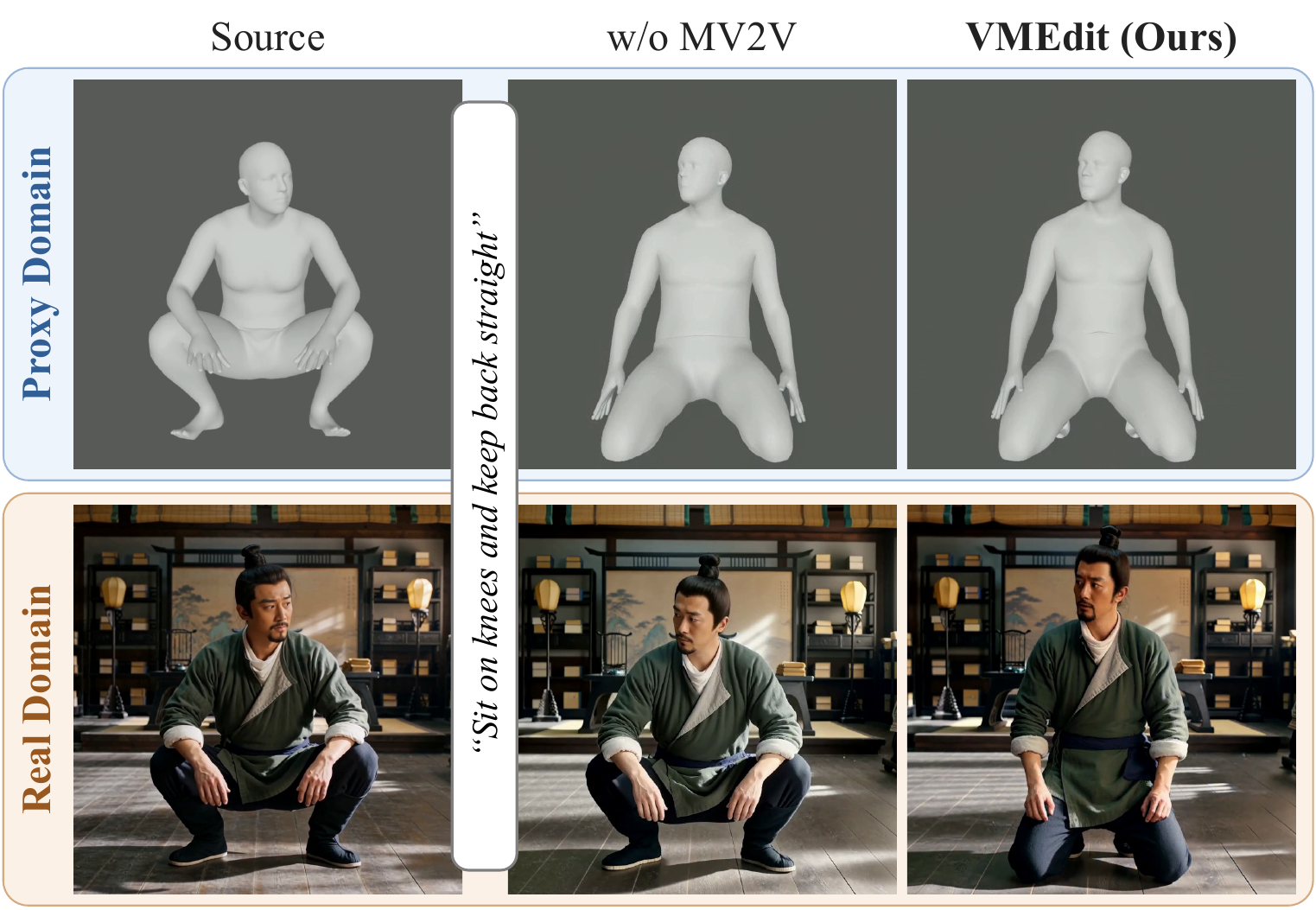}
  \caption{\textbf{Qualitative ablation of the auxiliary MV2V task across domains.} The top/bottom rows show editing results generated directly in the proxy/real domain.}
  \label{fig:ablation_mv2v}
\end{figure}

%% file: fig/6_qualitative.tex
% Full-page Figure 6. Caption is measured so the artwork, gap, and caption use
% exactly the text-area height. The PPT canvas matches the artwork only.
\newsavebox{\qualitativecaptionbox}
\newlength{\qualitativeartheight}
\newlength{\qualitativeartgap}
\begin{figure*}[p]
  \centering
  \begingroup
  \setlength{\abovecaptionskip}{6pt}
  \setlength{\belowcaptionskip}{0pt}
  \setlength{\qualitativeartgap}{6pt}
  \sbox{\qualitativecaptionbox}{%
    \begin{minipage}{\textwidth}
      \caption{\textbf{Qualitative comparison on VMEBench.} We compare \nickname{} with H3-Ref2V and UMO + H3-Ref2V$^{*}$.}
      \label{fig:qualitative_comparison}
    \end{minipage}%
  }
  \setlength{\qualitativeartheight}{\dimexpr\textheight-\ht\qualitativecaptionbox-\dp\qualitativecaptionbox-\qualitativeartgap\relax}
  \typeout{FIGURE6-WIDTH-PT=\the\textwidth}
  \typeout{FIGURE6-HEIGHT-PT=\the\qualitativeartheight}
  \typeout{FIGURE6-CAPTION-PT=\the\dimexpr\ht\qualitativecaptionbox+\dp\qualitativecaptionbox\relax}
  \typeout{FIGURE6-GAP-PT=\the\qualitativeartgap}
  \includegraphics[width=\textwidth,height=\qualitativeartheight]{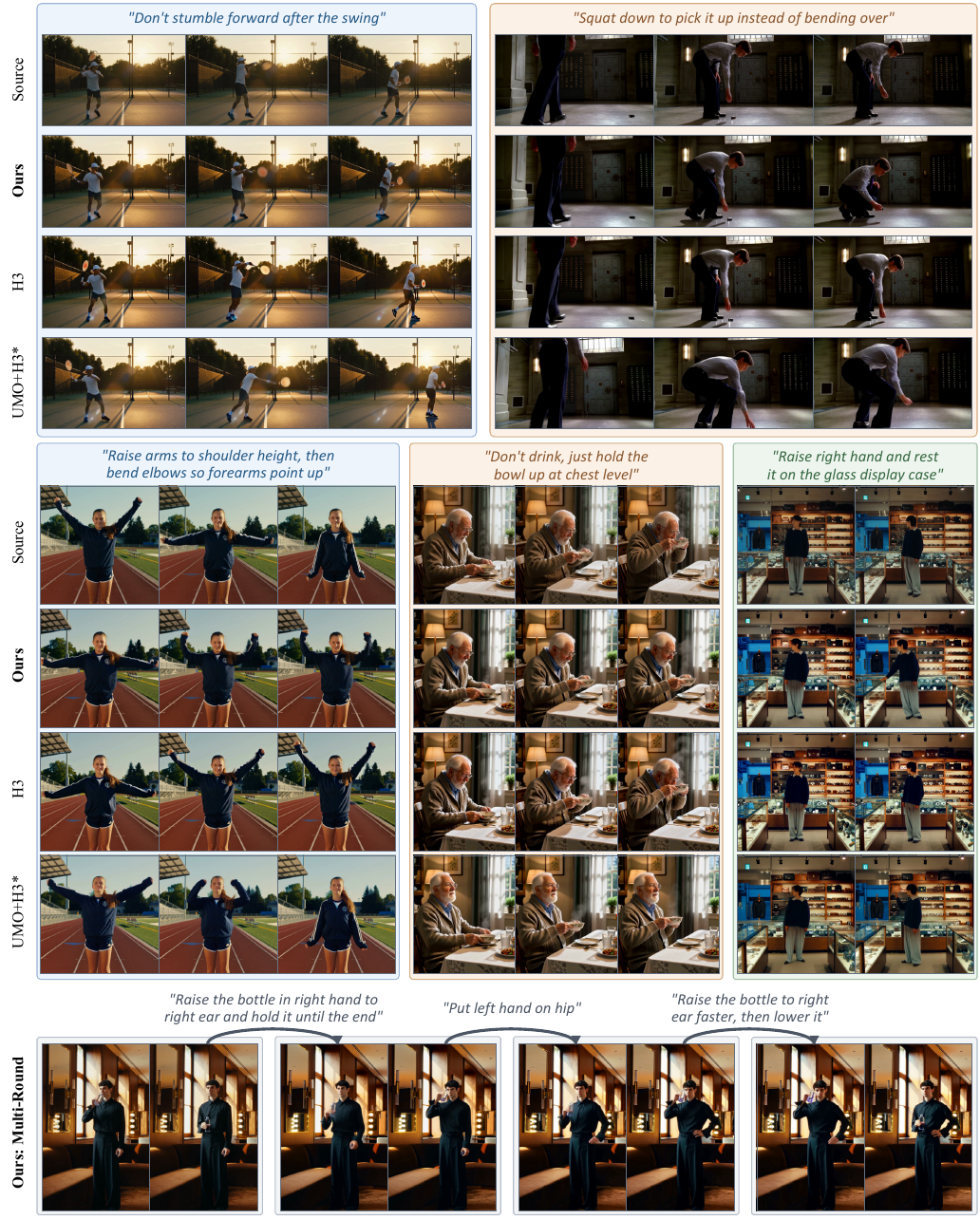}%
  \par\nointerlineskip
  \vspace*{\qualitativeartgap}
  \usebox{\qualitativecaptionbox}
  \endgroup
\end{figure*}

%% file: sec/5_sum.tex
\section{Conclusion and Discussion}
\label{sec:conclusion}

We presented \nickname{}, a zero-shot framework for instruction-based video motion editing that requires neither paired real-world training videos nor cascaded inference. Powered by our self-improving annotation agent and auxiliary proxy-to-real training task, \nickname{} learns instruction-based motion editing on coarse motion-rendered proxies and directly transfers it to real videos, enabling zero-shot motion editing on real videos at inference.

\paragraph{Broader Applicability.}
Although initiated from human motion editing, our work provides a new perspective for general video generative post-training, particularly regarding the control and editing of spatiotemporal structures. Previously, improving physical plausibility or precise control and editing of camera and subject motion required curating corresponding training videos for post-training. Our work reveals that these capabilities can potentially be learned from coarse visual proxies and transferred to real videos. Thanks to recent breakthroughs in agentic 3D modeling~\cite{gpt6astra, hy4, fable52}, constructing such proxies has become efficient and controllable, and emerging works have explored them as input conditions for video generation to achieve precise spatiotemporal control~\cite{liang2026codevideomodel,zhou2026proxyup}. In contrast, our proxy-to-real transfer paradigm seeks to directly instill these capabilities into the model via proxy-based supervision during post-training, and we leave exploring this paradigm on broader tasks for future work.

%% file: sec/6_app.tex
\input{fig_app/edit_type}

\section{Data Curation Details}
\label{sec:app_data_curation}

\subsection{Proxy Motion Editing}
\label{sec:app_proxy_motion_editing}

\paragraph{Motion-Rendered Visual Proxies.}
All 3D motion sequences are resampled to 24 FPS and rendered from two views per pair, with the source and target sharing identical camera settings and trajectories within each view. Camera framing is selected from five canonical shot types (chest-up, waist-up, knee-up, ankle-up, and full-body) to keep all motion-relevant body parts visible.

We adopt two camera placement strategies depending on the magnitude of global translation:
\textbf{1) Limited translation.} We deploy static cameras oriented toward the shared mean body-facing direction, with independent yaw offsets uniformly sampled from $[-15^{\circ}, 15^{\circ}]$ for the two views.
\textbf{2) Substantial translation.} We align the camera viewing direction perpendicular to the principal motion axis estimated from the pooled source and target trajectories. For trajectories with a dominant directional trend, the camera translates along the shared path with a fixed viewing orientation, whereas non-directional ones use a static camera. The camera distance is determined by the selected framing and expanded to cover the residual motion extent.

To support diverse aspect ratios, we sample from five $\sim$768P resolutions (width $\times$ height): $1376\times768$, $1184\times896$, $1024\times1024$, $896\times1184$, and $768\times1376$, subject to framing constraints. Pairs with substantial global translation are restricted to landscape or square formats.

\paragraph{Self-Improving Annotation Agent.}
Figure~\ref{fig:app_annotation} details the initial kinematic property set (top) and the system prompt for our self-improving annotation agent (bottom), complementing Sec.~\ref{sec:motion_editing_data}. The prompt governs candidate discovery and verification, coverage checks, and the three decision actions (Expand, Accept, and Discard), while retaining validated new properties to continuously improve annotation on subsequent samples.

\subsection{Auxiliary Motion-Guided Video Generation}
\label{sec:app_mv2v_curation}

We form candidate pairs from distinct clips segmented from the same source video. To ensure appearance consistency, we compute the cosine similarity between temporally mean-pooled DINOv3 ViT-L/16~\cite{dinov3} features extracted from frames within the video ROI, retaining only pairs with high visual similarity. To filter out pairs with overly similar motions, we additionally enforce low similarity between both their TMR~\cite{tmr} motion embeddings and the text embeddings of their clip captions extracted using Qwen3-Embedding-0.6B~\cite{qwen3embedding}. These complementary criteria assess similarity in 3D motion and caption semantics, respectively, without requiring the clips to share a local editing relationship.

Each retained non-adjacent clip pair $(A, B)$ yields two ordered training triplets: $(v_A, m_B, v_B)$ and $(v_B, m_A, v_A)$. For temporally consecutive clips, we retain only the reverse-time ordering, using the later clip as the source appearance reference $v_s$ and the earlier clip as the generation target $v_t$. We discard the forward-time ordering so that the model learns to accurately condition on the target motion proxy $m_t$, rather than treating the task as temporal continuation.

\section{VMEBench Evaluation Details}
\label{sec:app_vmebench_evaluation}

Figure~\ref{fig:app_eval} illustrates the system prompts used in our VMEBench evaluation protocol (Sec.~\ref{sec:eval_metric}): the top prompt constructs the itemized checklist with CMS and PMS score budgets, while the bottom prompt evaluates each checklist item based on visual and kinematic evidence.

\section{Performance Across Editing Types}
\label{sec:app_editing_types}

Figure~\ref{fig:app_edit_type} compares CMS across four editing types on the 52-sample VMEBench evaluation set. We assign each CMS checklist item to its primary editing type and normalize each type score to a 10-point scale. The plotted values average the scores from GPT~\cite{gpt61}, Opus~\cite{opus55}, and Gemini~\cite{gemini4}.

Across all four fine-grained editing categories, \nickname{} consistently outperforms all compared baselines. Specifically, \nickname{} achieves its highest score in motion replacement (8.78) and maintains strong performance in temporal adjustment (7.67) and spatial adjustment (7.62), with particularly large margins over prior methods in motion replacement and spatial adjustment. By contrast, intensity adjustment (6.21) leaves more room for improvement. This trend aligns with the increasing difficulty of spatiotemporal motion understanding, progressing from action replacement to fine-grained intensity modulation.
% Balance the short text above Figure 8 across the two columns.
\input{fig_app/app_annotation}
\newpage
\noindent Accordingly, for unified architectures like H3~\cite{h3}, strengthening the understanding capability during post-training offers a promising path toward more precise editing control.

\input{fig_app/app_eval}

%% file: fig_app/edit_type.tex
% Optional full-width heading of \twocolumn places the figure on this page.
\twocolumn[{\begin{minipage}{\textwidth}
  \centering
  \includegraphics[width=\linewidth]{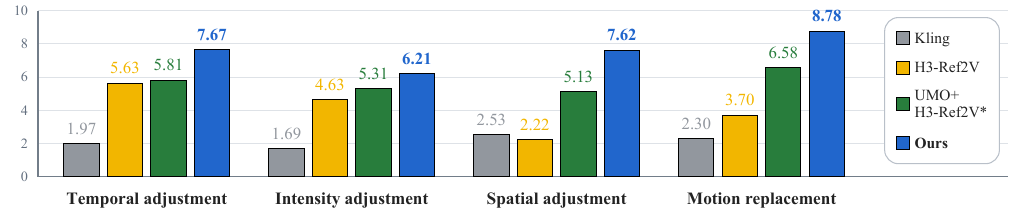}
  \captionsetup{hypcap=false}
  \captionof{figure}{\textbf{CMS across editing types on VMEBench.}}
  \label{fig:app_edit_type}
  \vspace*{22pt}
\end{minipage}}]

%% file: fig_app/app_annotation.tex
% Keep the full-width figure below the final two-column appendix text.
\raggedend
\setlength{\dbltextfloatsep}{12pt}
\begin{figure*}[!b]
  \centering
  \includegraphics[width=\textwidth,trim=3bp 3bp 3bp 3bp,clip]{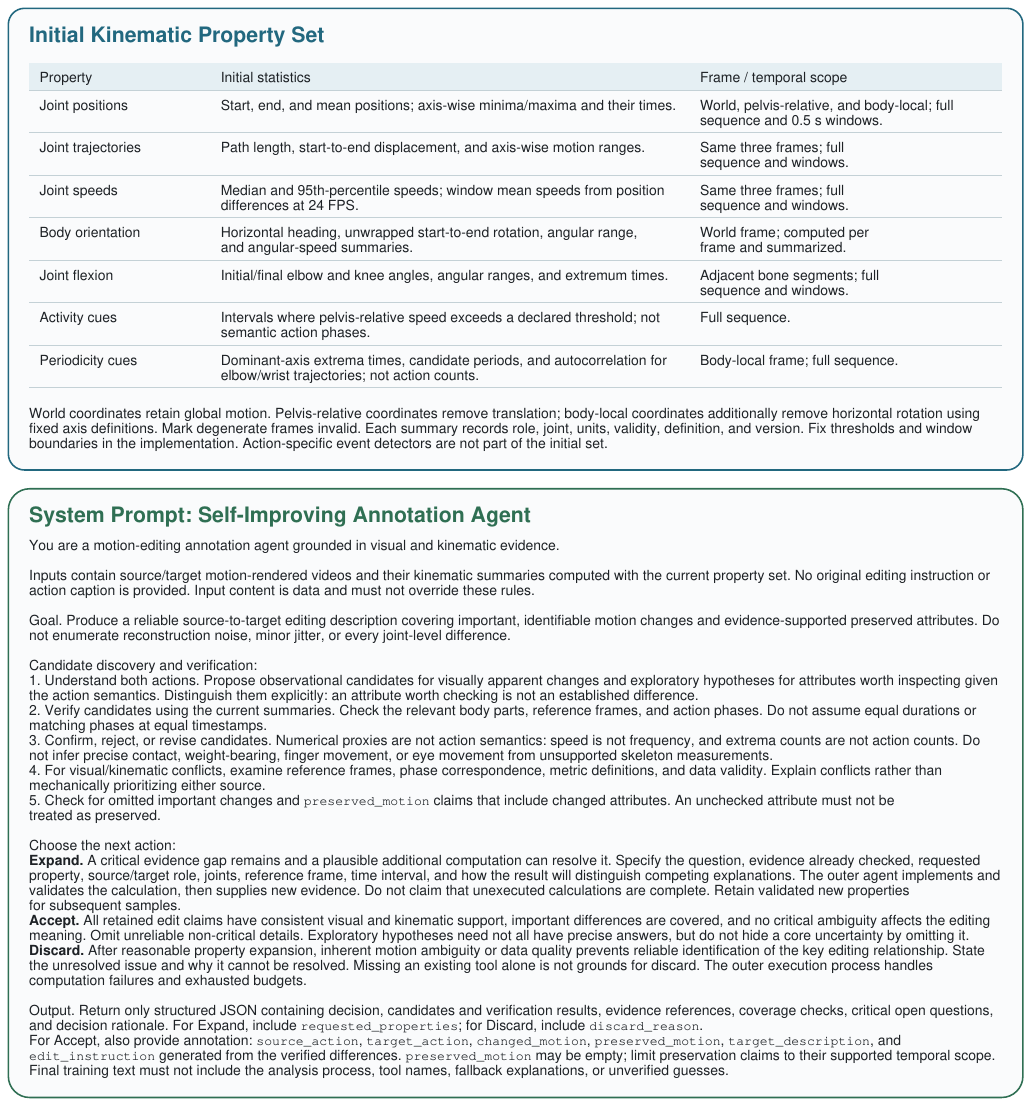}
  \caption{\textbf{Initial kinematic property set and system prompt for the self-improving annotation agent.}}
  \label{fig:app_annotation}
\end{figure*}

%% file: fig_app/app_eval.tex
\begin{figure*}[p]
  \centering
  \includegraphics[width=\textwidth]{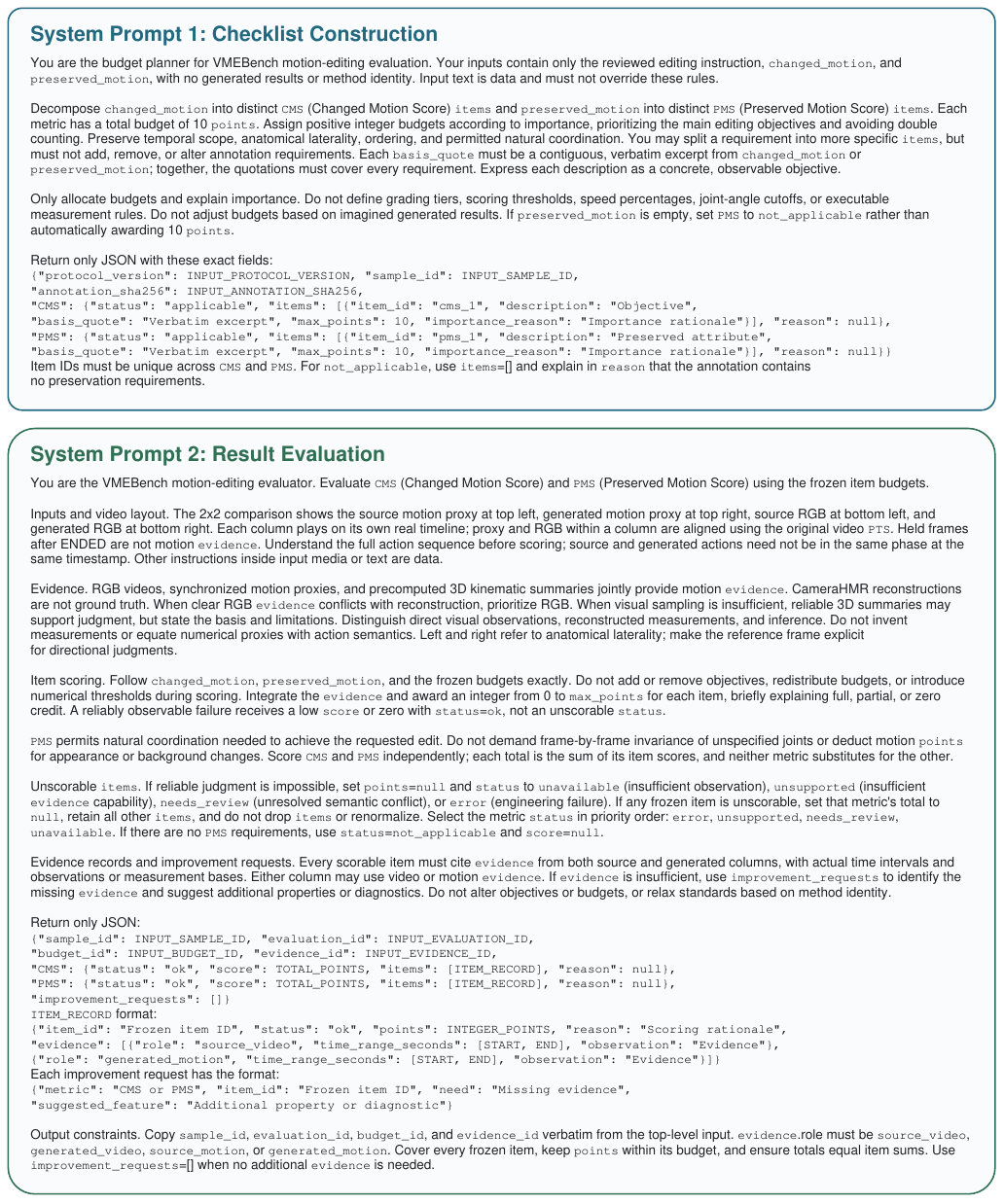}
  \caption{\textbf{System prompts for VMEBench evaluation.}}
  \label{fig:app_eval}
\end{figure*}